\documentclass[letterpaper, 10 pt, journal, twoside]{ieeetran}  %

\IEEEoverridecommandlockouts                              %

\newcommand{\subparagraph}{}

\usepackage{graphicx}
\usepackage{amsmath} %
\usepackage{amssymb} %
\usepackage{subfig}
\usepackage{float}
\usepackage{titlesec}

\usepackage{booktabs}
\usepackage{multirow}

\makeatletter
\let\NAT@parse\undefined
\makeatother

\usepackage[breaklinks=true,colorlinks,bookmarks=false]{hyperref}
\usepackage[noabbrev,capitalise]{cleveref}

\usepackage{url}
\usepackage{color}

\newcommand\Tstrut{\rule{0pt}{2.6ex}}       %
\titlespacing*{\section}{0pt}{1.5mm}{1.5mm}
\titlespacing*{\subsection}{0pt}{.75mm}{.75mm}
\titlespacing*{\subsubsection}{0pt}{1mm}{1mm}

\renewcommand{\baselinestretch}{0.93}

\title{Multimodal Deep Generative Models for Trajectory Prediction: A Conditional Variational Autoencoder Approach
}

\author{Boris Ivanovic$^{1*}$, Karen Leung$^{1*}$, Edward Schmerling$^2$, Marco Pavone$^1$%
\thanks{Manuscript received: July 1, 2020; Revised October 13, 2020; Accepted November 6, 2020.}%
\thanks{This paper was recommended for publication by Editor Tamim Asfour upon evaluation of the Associate Editor and Reviewers' comments.
Toyota Research Institute (TRI) provided funding to assist the authors with their research but this article solely reflects the opinions and conclusions of its authors and not TRI or any other Toyota entity. We thank Matteo Zallio for his help in visually communicating our ideas.}%
\thanks{$^*$Denotes equal contribution.}%
\thanks{$^1$Department of Aeronautics and Astronautics, Stanford University.
        {\tt\small  \{borisi, karenl7, pavone\}@stanford.edu}}%
\thanks{$^2$Institute for Computational \& Mathematical Engineering, Stanford University. {\tt\small  schmrlng@stanford.edu}}%
\thanks{Digital Object Identifier (DOI): see top of this page.}
}

\begin{document}

\maketitle
\begin{abstract}
Human behavior prediction models enable robots to anticipate how humans may react to their actions, and hence are instrumental to devising safe and proactive robot planning algorithms.
However, modeling complex interaction dynamics and capturing the possibility of many possible outcomes in such interactive settings is very challenging, which has recently prompted the study of several different approaches. In this work, we provide a self-contained tutorial on a conditional variational autoencoder (CVAE) approach to human behavior prediction which, at its core, can produce a multimodal probability distribution over future human trajectories conditioned on past interactions and candidate robot future actions.
Specifically, the goals of this tutorial paper are to review and build a taxonomy of state-of-the-art methods in human behavior prediction, from physics-based to purely data-driven methods,
provide a rigorous yet easily accessible description of a data-driven, CVAE-based approach, highlight important design characteristics that make this an attractive model to use in the context of model-based planning for human-robot interactions,
and provide important design considerations when using this class of models.
\end{abstract}


\section{Introduction}\label{sec:intro}

\begin{figure}[t]
    \centering
    \includegraphics[width=\linewidth]{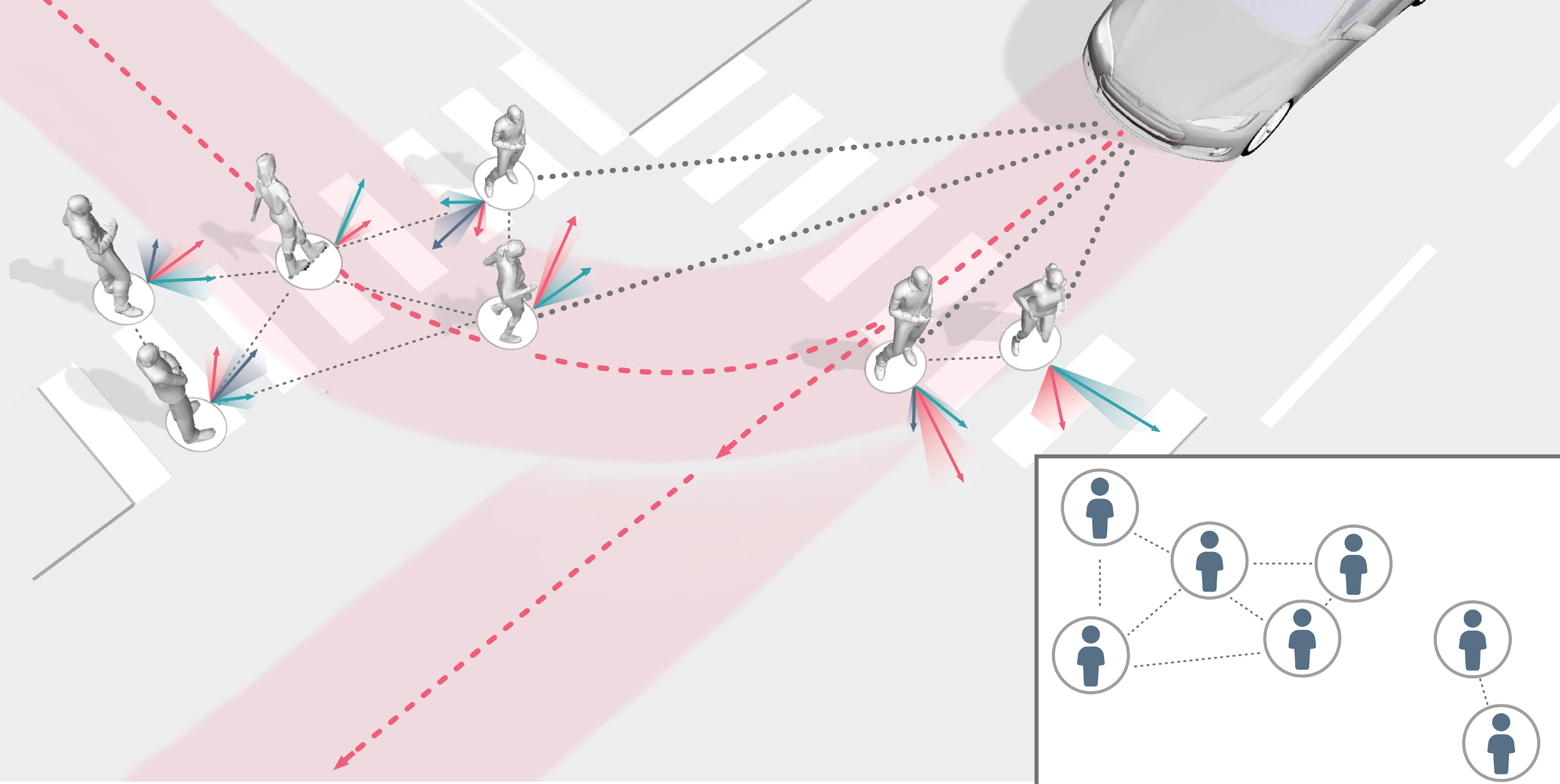}
    \caption{
    There are many different ways an interaction (e.g., pedestrians crossing the road) may evolve. 
    For safe human-robot interactions, robots (e.g., autonomous car) need to reason about the possibility of multiple outcomes (denoted by the colored shaded arrows), and understand how their actions influence the actions of others. Inset: Graphical representation of the interaction.
    }
    \label{fig:hero}
\end{figure}

\IEEEPARstart{H}{uman} behavior is inconsistent across populations, settings, and even different instants, with all other factors equal---addressing this inherent uncertainty is one of the fundamental challenges in human-robot interaction (HRI).
Even when a human’s broader intent is known, there are often multiple distinct courses of action that person may pursue to accomplish their goals. 
For example, in Figure~\ref{fig:hero}, a pedestrian crossing a road may pass to the left or right of an oncoming pedestrian; reasoning about the situation cannot be simplified to the ``average'' case, i.e., the pedestrians colliding.
To an observer, the choice of mode may seem to have a random component, yet also depend on the evolution of the human’s surroundings. %
Imbuing a robot with the ability to take into consideration the full breadth of possibilities in how humans may respond to its actions is a key component of enabling anticipatory and proactive robot decision-making policies which can result in safer and more efficient interactions.

For the goal of creating robots that interact intelligently with human counterparts, 
observing data from human-human interactions has provided valuable insight into modeling interaction dynamics (see \cite{RudenkoPalmieriEtAl2019} for an extensive survey). A robot may reason about human actions, and corresponding likelihoods, based on how it has seen humans behave in similar settings.
To implement a robot's control policy, model-free methods tackle this problem in an end-to-end fashion---human behavior predictions are implicitly encoded in the robot's policy which is learned directly from data. On the other hand, model-based methods decouple the model learning and policy construction---a probabilistic understanding of the interaction dynamics is used as a basis for policy construction.
By decoupling action/reaction prediction from policy construction, model-based approaches often afford a degree of transparency in a planner’s decision making that is typically unavailable in model-free approaches.
In this paper, we take on a model-based approach to HRI and focus on learning a model of human behaviors, or more specifically, distributions over future human behaviors (e.g., trajectories).

Within model-based methods for HRI, there are many existing approaches to modeling human behaviors, and they can broadly be categorized as ontological or phenomenological.
To contextualize our work against other methods, we will build a taxonomy of different types of ontological and phenomenological state-of-the-art methods in this field.  
We note that these methods could be categorized differently across other dimensions (e.g., whether the model produces probabilistic or deterministic predictions).
At a high-level, ontological approaches (sometimes referred to as ``theory of mind'') postulate a core underlying structure about an agent's behavior
and build a mathematical model upon it. For instance, they may craft a set of rules that agents are required to follow, or an analytic model describing an agent's internal decision-making scheme.
In contrast, phenomenological approaches do not make such strong modeling assumptions, and instead rely on a wealth of data to model agent behaviors without explicitly reasoning about underlying motivations.

We approach this problem phenomenologically, and in particular, focus on using a Conditional Variational Autoencoder (CVAE) \cite{SohnLeeEtAl2015} to learn a human behavior prediction model well-suited for
model-based planning and control \cite{SchmerlingLeungEtAl2018}.
We seek to explicitly characterize the multimodal uncertainty in human actions at each time step conditioned on interaction history as well as future robot action choices. Conditioning on interaction history allows a robot to reason about hidden factors like experience, mood, or engagement level that may influence the distribution, and conditioning on the robot's next action choices takes into account response dynamics.
Due especially to this latter capability, conditional behavior prediction models have achieved great success when applied to planning in interactive scenarios,
however, a ubiquitous problem with such models is that they may not be able to distinguish between correlation and causation when learned from offline data. Though the focus of this paper is on a specific methodology for conditional behavior prediction, we briefly note that mitigations for this issue are an area of active research and include data gathering strategies to disambiguate causal confusion \cite{HaanJayaramanEtAl2019}, and enacting safety-preserving controllers at a level beneath the planner which provide a fallback in the case of incorrect inference \cite{LeungSchmerlingEtAl2019}.

\emph{Goals of this paper}: The main goal of this paper is to provide a self-contained tutorial on the CVAE-based human trajectory prediction model proposed and developed in \cite{SchmerlingLeungEtAl2018,IvanovicSchmerlingEtAl2018,IvanovicPavone2019}, and \cite{SalzmannIvanovicEtAl2020}. Before we delve into the details of our approach, we build a taxonomy of state-of-the-art methods for human behavior prediction in interactive settings in order to provide insight into the problem setups and system design goals for which our work is best suited.
Thus, the contributions of this paper are five-fold: we
(i) provide a concise taxonomy of ontological and phenomenological methods for human behavior prediction for interactive settings with discussion tailored to motivating our approach (Section~\ref{sec:related work}), 
(ii) introduce CVAEs in a self-contained manner and detail the proposed neural network architecture for human trajectory prediction (Section~\ref{sec:cvae}), 
(iii) demonstrate the benefits of the model with a focus on its scalability to multi-agent settings, use of hetereogeneous data, and ability to produce an analytic representation of the output trajectory distribution rooted with a dynamics model (Section~\ref{sec:multiagent} and \ref{sec:dyn_hetero_data}), and
(iv) compare the performance of such an approach against other state-of-the-art phenomenological approaches as well as discuss important implementation considerations for practitioners using this model (Section~\ref{sec:expts}).

\section{Related Work}\label{sec:related work}
Methods for predicting human behaviors can be classified as either ontological or phenomenological.
Ontological models make assumptions about an agent's dynamics or motivation.
One direction is to make assumptions about the underlying physics that govern the system and then derive a state-space model via first principles.
For instance, the Social Forces model \cite{HelbingMolnar1995} formulates the interaction dynamics by making assumptions on the attractive and repulsive forces between agents.
Similarly, the Intelligent Driver Model (IDM) \cite{TreiberHenneckeEtAl2000} derives a continuous-time car-following differential equation model.
Due to the simplicity of these models, they are very useful in simulating large-scale interactions, such as crowd dynamics \cite{HelbingFarkasEtAl2000} or traffic flow \cite{TreiberHenneckeEtAl2000b}.
Despite these methods capturing the coupling between agents, they are fundamentally unimodal representations of the interaction (i.e., do not account for the possibility of multiple distinct futures) and do not utilize knowledge of past interactions. 

Rather than formulating the interaction dynamics explicitly, we can instead make assumptions about a human's internal decision-making process. 
Game theoretic approaches model the interaction dynamics by making assumptions on whether the other agent is cooperative \cite{NikolaidisNathEtAl2017} or adversarial \cite{WangWangEtAl2019} and leverage this information for robot planning. For socially-aware robot navigation, \cite{NarayananManogharEtAl2020,RandhavaneBeraEtAl2019} infer a human's emotion or dominance and use that to inform their robot planner.
A popular approach is to model humans as optimal planners and represent their motivations at each time step as a state/action dependent reward (equivalently, negative cost) function.
Maximizing this function, e.g., by following its gradients to select
next actions, may be thought of as a computational proxy
for human decision-making processes. 

Inverse Reinforcement Learning (IRL) \cite{NgRussell2000,AbbeelNg2004} is a generalization of this idea whereby a parameterized family of reward functions is fit to a dataset of human state-action demonstrations.
The reward function is typically represented as a linear combination of possibly nonlinear features $r(x,u) = \theta^T\varphi(x,u)$, where the weight parameters $\theta$ are fit to minimize a measure of error between the actions that optimize $r$ and
the true human actions. 
One of the typical strengths of IRL is its interpretability, both in terms of the ability to include handcrafted features, as well as what learned linear weights reveal regarding feature importance.
Maximum entropy (MaxEnt) IRL \cite{ZiebartMaasEtAl2008} applies this principle in a probabilistic fashion; the probability distribution over human actions is proportional to the exponential of the reward $p(u) \propto \exp(r(x,u))$.
This framework has been employed to model human behaviors in the context of driving \cite{SadighSastryEtAl2016c} and social navigation \cite{KretzschmarSpiesEtAl2016} and then used to inform a robot's planning strategy.
In theory, with sufficiently complex and numerous features in the reward function, MaxEnt IRL could approximate any (including multimodal) distributions arbitrarily well, making this an attractive candidate for our application of HRI. However, there are two main drawbacks with typical applications of MaxEnt IRL that prompt us to consider an alternative approach.
First, though the learned distribution may be multimodal, if it is represented as an unnormalized log-probability density function (i.e., $r(x,u)$), there may not be a computationally tractable way to reason about this multimodality when planning (e.g., by sampling).
Previous work has relied on search for explicit mode enumeration \cite{KretzschmarSpiesEtAl2016}, or in the case of \cite{SadighSastryEtAl2016c}, which develops a unified and tractable framework for MaxEnt IRL-based prediction and policy construction for intelligent vehicles, the resulting policy employs gradient-based local optimization which ultimately results in a unimodal assumption on interaction outcome despite learning a nominally multimodal distribution.
Second, IRL is typically applied to learn importance weights for a handful of human-interpretable features. Using more complex, possibly deep-learned, features to increase expressivity of the model removes one of the key benefits of IRL, and instead promotes the use of phenomenological methods. For example, though this is not a fundamental limitation of IRL, to maximize interpretability existing work has often made a Markovian assumption in constructing features that depend only on the current state \cite{SadighSastryEtAl2016c}, and thus do not capture interaction history when reasoning about future behavior.
In general, reward-based approaches can be effective in settings with limited data as there are only a few parameters to learn, and can transfer to new and unseen tasks \cite{ChoudhurySwamyEtAl2019}. However, in the presence of large amounts of data and with the desire to condition on interaction history, it is natural to consider phenomenological approaches.

Phenomenological approaches
are methods that do not make inherent assumptions about the structure of the interaction dynamics and/or agents’ decision-making process. Instead, they rely on powerful modeling techniques and a wealth of observation data to infer and replicate the complex interactions. 
Recently, there have been a plethora of deep learning-based regression models for predicting future human trajectories (e.g., \cite{AlahiGoelEtAl2016,JainZamirEtAl2016}) following the success of Long Short-Term Memory (LSTM) networks \cite{HochreiterSchmidhuber1997}, a purpose-built deep learning architecture for modeling temporal sequence data.
However, such methods only produce a single deterministic trajectory output and therefore neglect to capture the uncertainty inherent in human behaviors. 
Safety-critical systems need to reason about many possible future outcomes to guard against worst-case scenarios, ideally with the likelihoods of each occurring, to enable safe decision-making. 
As a result, there has been recent interest in methods that simultaneously forecast multiple possible futures, or produce a distribution over possible future outcomes.

Due to recent advancements in generative modeling, 
\cite{SohnLeeEtAl2015,GoodfellowPouget-AbadieEtAl2014}, 
there has been a paradigm shift from deterministic regressors to generative models, i.e., models that produce a distribution over possible future behaviors. In particular, deep generative methods (neural-network based models that learn an approximation of the true underlying probability distribution from which the the dataset was sampled from) have emerged as state-of-the-art approaches.
There are two main deep generative methods that dominate the field, (Conditional) Generative Adversarial Networks ((C)GANs) \cite{GoodfellowPouget-AbadieEtAl2014,MirzaOsindero2014}, and (Conditional) Variational Autoencoders ((C)VAEs) \cite{KingmaWelling2013, SohnLeeEtAl2015}.
Both these methods have been widely used in the context of future human trajectory prediction in interactive settings (e.g., \cite{GuptaJohnsonEtAl2018,LeeChoiEtAl2017, DeoTrivedi2018,KosarajuSadeghianEtAl2019}).
GANs are composed of a generator and discriminator network---to produce realistic outputs, the generator outputs samples which are then ``judged'' by the discriminator. 
Although GAN-based models show promising results, there are two main limitations. First, GAN learning often suffers from mode collapse, a phenomenon where the model converges to the mode of the distribution and is unable to capture and produce diverse outputs \cite{SalimansGoodfellowEtAl2016}. 
This is incompatible with safety-critical applications where it is important to capture rare yet potentially catastrophic outcomes.
Second, GANs are notoriously difficult to train because the conflict between the generator and discriminator causes instability in the training process \cite{ArjovskyBottou2017,ArjovskyChintalaEtAl2017}. 
Additionally, despite offering flexibility in the definition of the objective function, GANs fundamentally output an empirical distribution of samples which may limit the types of model-based planners/controllers that can be used (e.g., planners that rely on a parameterized distribution). 

Alternatively, (C)VAEs take on a variational Bayesian approach; they learn an approximation of the true underlying probability distribution by distilling latent attributes as probability distributions and then ``decode'' samples from the latent distribution to produce desired outputs.
In contrast to GANs, (C)VAEs optimize the likelihood over all examples in the training set, meaning all the modes of the distribution are considered, and are less likely to suffer problems of mode collapse and lack of diversity seen with GANs.
Additionally, (C)VAEs can produce either empirical samples from the distribution, or an analytical representation of the distribution, making them potentially more versatile than GANs in the context of model-based planning and control. 

There are thus many considerations when selecting a method to model interaction dynamics and perform human behavior prediction. In HRI settings with large amounts of data available, and the need for high expressivity to capture interaction nuances and multimodal distribution coverage over the output space, we focus the remainder of this work on using CVAEs for human trajectory prediction.

\section{The Conditional Variational Autoencoder for Interaction-aware Behavior Prediction}\label{sec:cvae}

We describe a general CVAE model and apply it in the context of human behavior prediction. We highlight the core characteristics of our proposed CVAE trajectory prediction model and illustrate them with a traffic-weaving case-study.

\subsection{Conditional Variational Autoencoder (CVAE)}

\begin{figure*}[t]
    \centering
    \subfloat[Graphical model of a CVAE.]{
    \label{fig:CVAE graphical model}
    \includegraphics[width=0.3\textwidth]{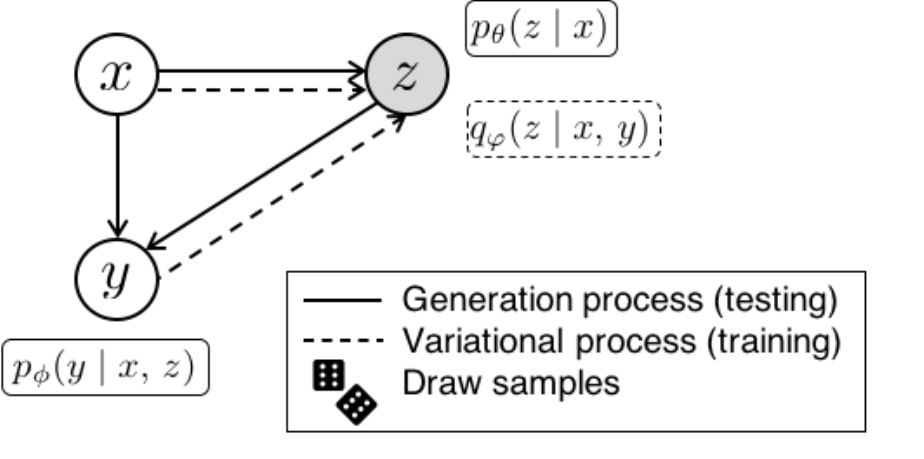}} $\:\:$ \subfloat[Sequence-to-sequence CVAE architecture for human behavior prediction.]{
     \label{fig:proposed seq-to-seq cvae architecture}
    \includegraphics[width=0.63\textwidth]{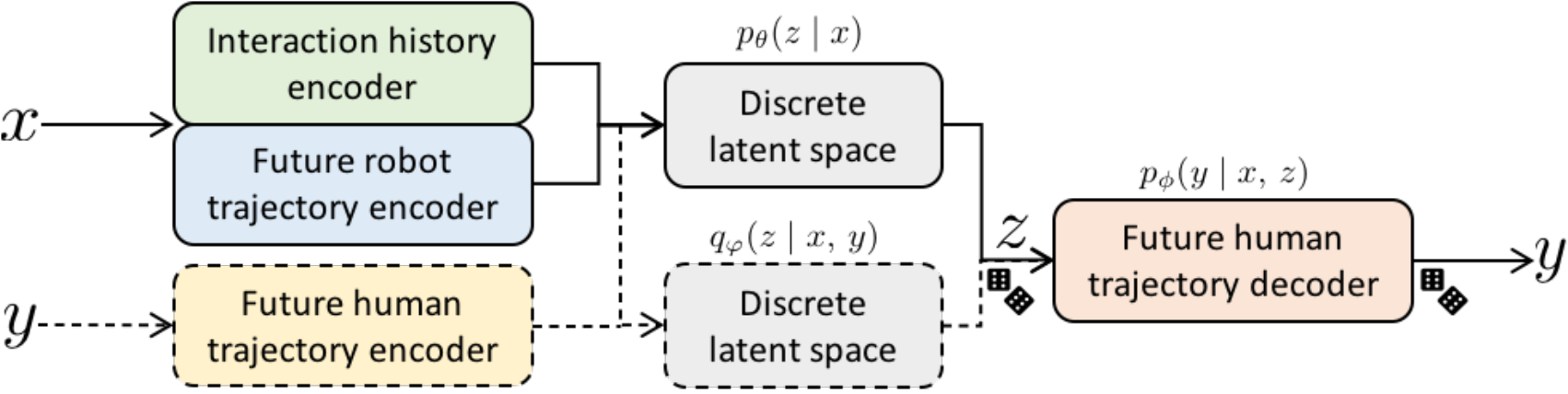}}
\caption{A graphical model of a CVAE, and a neural network architecture of a CVAE for human behavior prediction. Solid lines represent components for the generation process (during testing), and dashed lines represent components used for variational inference (during training).}
\label{fig:CVAE graphical model and neural architecture}
\end{figure*}

Given a dataset $\mathcal{D} = \lbrace (x_i, y_i) \rbrace_{i=1}^N$, the goal of \emph{conditional} generative modeling is to fit a model of the conditional probability distribution $p(y\mid x)$, which may be used for downstream applications such as inference (i.e., calculating the likelihood of observing a particular sample $y$ given $x$), or to generate new samples $y$ given $x$.
In this work we consider parametric models, whereby we consider $p(y\mid x)$ within a family of distributions defined by a fixed set of parameters, which we fit to the dataset with the objective of maximizing the likelihood of the observed data. Due to their expressivity, neural networks are often used to represent complex and high-dimensional distributions.
A CVAE \cite{SohnLeeEtAl2015} is a \emph{latent} conditional generative model. The goal is still to approximate $p(y\mid x)$, but before outputting $p(y\mid x)$ the model first projects the inputs into a lower dimensional space, called the \emph{latent space}, which acts as a bottleneck to encourage the model to uncover salient features with the intended purposes of improving performance, and potentially aiding in interpretability.
Figure~\ref{fig:CVAE graphical model} illustrates the graphical model of a CVAE. 
An encoder, parameterized by $\theta$, takes the input $x$ and produces a distribution $p_\theta(z \mid x)$ where $z$ is a latent variable that can be continuous or discrete \cite{JangGuEtAl2017,MaddisonMnihEtAl2017}.\footnote{For this work, we focus on a discrete latent space, but note the following equations still apply by replacing the summation with an integral.}
A decoder, parameterized by $\phi$, uses $x$ and samples from $p_\theta(z \mid x)$ to produce $p_\phi(y \mid x, \, z)$. In practice, the encoder and decoder are neural networks. The latent variable $z$ is then marginalized out to obtain $p(y\mid x)$,
\begin{equation}
    p(y \mid x) = \sum_z p_\phi(y \mid x, \, z) p_\theta(z \mid x).
    \label{eqn:conditional probability marginalizing z}
\end{equation}
To efficiently perform the marginalization in \eqref{eqn:conditional probability marginalizing z}, we desire values of $z$ that are likely to have produced $y$, otherwise $p_\theta(z\mid x) \approx 0$ and will contribute almost nothing to $p(y \mid x)$.\footnote{We note that if the size of the discrete latent space is small, we can tractably compute the summation in \eqref{eqn:conditional probability marginalizing z} exactly.} 
To this end, we perform \emph{importance sampling} by instead sampling from $q(z \mid x, \, y)$, a proposal distribution, which will help us select values of $z$ that are likely to have produced $y$. 
Since we are free to choose $q(z \mid x ,\, y)$, we parameterize it (often as a neural network) with $\varphi$, denoted by $q_\varphi(z \mid x ,\, y)$.
We can rewrite \eqref{eqn:conditional probability marginalizing z} by multiplying and dividing by the proposal distribution, and using the definition of expectation,
\begin{equation*}
\begin{aligned}
p(y \mid x) =& \: \sum_z \frac{p_\phi(y \mid x, \, z) p_\theta(z \mid x)}{q_\varphi(z \mid x, \, y)} q_\varphi(z \mid x, \, y)\\
=& \:\mathbb{E}_{q_\varphi(z \mid x, \, y)} \left[\frac{p_\phi(y \mid x, \, z) p_\theta(z \mid x)}{q_\varphi(z \mid x, \, y)} \right]. \\
\end{aligned}
\end{equation*}
The goal is to fit parameters $\phi$, $\theta$, and $\varphi$ that maximize the log-likelihood of $p(y\mid x)$ over the dataset $\mathcal{D}$.
By taking the $\log$ of both sides, using Jensen's inequality, and rearranging the terms, the \emph{evidence lower-bound} (ELBO) is derived,
\begin{equation}
    \begin{aligned}
        \log p(y \mid x) \geq & \: \mathbb{E}_{q_\varphi(z \mid x, \, y)}\left[ \log p_\phi(y \mid x, \, z) \right] - \\
        & \:\qquad \quad \mathcal{D}_\mathrm{KL}\left[ q_\varphi(z \mid x,\,y) \, \| \,  p_\theta(z \mid x)\right]
    \end{aligned}
    \label{eqn:deriving the ELBO}
\end{equation}
where $\mathcal{D}_\mathrm{KL}(p \| q) = \mathbb{E}_{p(x)}\left[\log \frac{p(x)}{q(x)}\right]$ is the Kullback-Liebler divergence.
The ELBO is a lower bound on $\log p(y \mid x)$, the quantity that we are trying to maximize, but which is often intractable to compute directly by~\eqref{eqn:conditional probability marginalizing z}. Instead, we maximize the ELBO as a proxy.
By using the reparameterization trick \cite{KingmaWelling2013,JangGuEtAl2017,MaddisonMnihEtAl2017}, the ELBO is tractable to compute and can be optimized via stochastic gradient descent. The loss for a single training example $(x,y)$ is,
\begin{equation}
    \begin{aligned}
    \mathcal{L}(x,y) = &\, -\mathbb{E}_{q_\varphi(z \mid x, \, y)}\left[ \log p_\phi(y \mid x, \, z) \right] +\\ 
    & \qquad \mathcal{D}_\mathrm{KL}\left[ q_\varphi(z \mid x,\,y) \, \| \,  p_\theta(z \mid x)\right].
    \end{aligned}
    \label{eqn:ELBO loss function}
\end{equation}
During training, we minimize the Monte Carlo estimate of the expected loss over the training set.

\begin{figure*}[t]
    \centering
    \includegraphics[width=0.9\textwidth]{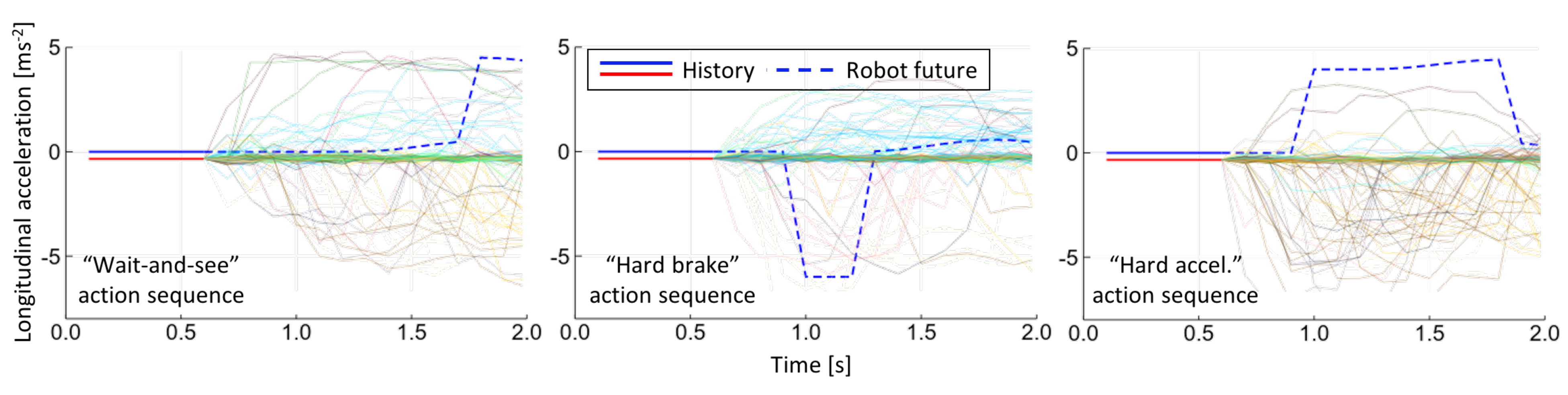}
    \caption{Predictions of future human action sequences depend on the future action sequence of the robot (blue dashed line). The different colors of future human action sequences correspond to different discrete latent variable instantiations (i.e., different modes in the multimodal output distribution). This figure has been adapted from \cite{SchmerlingLeungEtAl2018}.}
    \label{fig:traffic weaving interaction dynamics}
\end{figure*}

\subsection{Interaction-aware Human Behavior Prediction}
We are interested in learning a model that is able to predict future trajectories of intelligent agents (i.e., we assume these agents are humans, or human-controlled) interacting with other intelligent agents in the environment. Specifically, we desire a model that (i) is history dependent in order to capture behavioral tendencies or intent, (ii) accounts for the coupled interaction dynamics between all agents, (iii) produces a \emph{multimodal} distribution over future human trajectories because there are many different ways a human may behave in an interactive setting, and (iv) is well-suited for model-based planning since our ultimate goal is to design robots that can leverage these predictions to interact seamlessly with humans.
Our proposed sequence-to-sequence CVAE trajectory prediction architecture, shown in Figure~\ref{fig:proposed seq-to-seq cvae architecture}, is able to address these desiderata in the following ways.

To address (i) and (ii) above, the input conditioning variable $x$ consists of features representing \emph{interaction history}, a sequence of features from all agents (e.g., positions, velocities, actions) since the start of the interaction, and a future robot trajectory, a sequence of states and/or actions that the robot plans to follow over the planning horizon. We can additionally include other features that may be relevant to the application, such as a map of the environment, or camera images from the robot (see Section~\ref{sec:dyn_hetero_data}). The output $y$ is a sequence of future states/actions of all human agents that we are interested in. Since the output is conditioned in part on what the robot will do in the future, this model learns the coupled interaction dynamics. We discuss later in Section~\ref{sec:dyn_hetero_data} how we can integrate predicted action distributions to produce full dynamically-feasible trajectory predictions.

To address (iii), a multimodal distribution is constructed by using a \emph{discrete} latent space. Each latent vector instantiation of $z$ corresponds to a discrete mode (i.e., a mixture component), and its probability $p_\theta(z\mid x)$, is produced by the encoder (corresponding to the mixture weight). 
For example, one of the discrete modes might correspond to a human driver braking, while another might correspond to turning right. Note that enforcing semantic meaning to each latent value is by no means guaranteed, and is an area of active research \cite{AdelGhahramaniEtAl2018}. A continuous latent space could be used,
though in our work, we found a discrete latent space to be more effective. 
For a given mode, there are variations to how this behavior might occur (e.g., slightly different ways to turn right). To cater for these variations and account for dependencies in successive states or actions, the decoder outputs an autoregressive sequence of Gaussian mixture models (GMMs). We want to highlight that the use of GMMs here is not the main mechanism behind creating a multimodal distribution over trajectories; that is the role of the latent space.
At each time step of the prediction horizon, the decoder outputs GMM components describing the distribution of the output features, then a sample is drawn from the GMM and is used to generate a GMM at the next time step. Repeating this process will create a \emph{sample} drawn from $p(y\mid x)$. For the case with a single GMM component (i.e., a Gaussian), the mean and variance at each time step can be propagated instead of a sample, enabling an \emph{analytic} representation of the output distribution (see Section~\ref{sec:dyn_hetero_data}).

The flexibility in how the output distribution is represented addresses (iv); we can tailor the outputs to the needs of a model-based planner. Specifically, we can choose to describe the learned distribution empirically (i.e., output samples directly), or analytically (i.e., output parameters of the distribution).
Additionally, 
there are many options for how to construct the encoder and decoder. We primarily leverage recurrent neural networks (RNN) to process time-series data with potentially variable length without increasing the problem size.
As we will describe in Sections~\ref{sec:multiagent} and \ref{sec:dyn_hetero_data}, we can augment the model to consider spatio-temporal relationships between multiple agents and heterogeneous data inputs (e.g., state trajectories, images, and maps).

\subsection{Traffic-Weaving Case Study}\label{sec:two_agent_cvae}

We reproduce the traffic-weaving scenario studied in \cite{SchmerlingLeungEtAl2018} to illustrate the key characteristics of our approach.
In the traffic-weaving scenario, two cars initially side-by-side must swap lanes in a short amount of time and distance, emulating cars merging onto/off of a highway. This is a challenging negotiation due to the inherent multimodal uncertainty of who will pass whom. 
Before we begin, we make two remarks.
First, we use an LSTM for the encoder and decoder networks as we found this RNN architecture provided the best performance in terms of evaluation loss.
Second, we choose to predict future human \emph{action} sequences and use future robot \emph{action} sequences as inputs as this aligned with our case study. However, for other applications, states can be used instead of actions.

The interaction history is defined as the sequence of states and actions from both agents since the start of the interaction. We consider a future robot action sequence as an additional input; together with the interaction history this forms the conditioning variable $x$. The learned CVAE model defines a distribution $p_\theta(z\mid x)$ over the latent variable $z$ which is fed, with $x$, into the decoder $p_\phi(y \mid x, \, z)$ to produce predicted human action sequences $y$. The LSTM decoder produces GMM components describing a distribution over human actions at each time step; to produce the sequence $y$ an action is sampled from the GMM and is fed into back into the LSTM cell to produce the next action, and so forth.

In Figure~\ref{fig:traffic weaving interaction dynamics}, when the robot is deciding its next action to take, it can predict how the human may respond to each of its candidate future action sequences (dashed blue line). The different colors in the predictions (thin lines) showcase the different modes, i.e., discrete latent values $z$, in the output distribution. For instance, light blue trajectories correspond to the human speeding up, while dark yellow trajectories correspond to the human slowing down.
Given this interaction model, the robot can select next actions by searching over a set of possible future action sequences and selecting the one that results in the highest expected reward. This model-based planner was tested and validated in simulation \cite{SchmerlingLeungEtAl2018} and on a full-scale test vehicle \cite{LeungSchmerlingEtAl2019}.

\section{Scaling up to Multi-agent Interactions}\label{sec:multiagent}
In the real world, agents simultaneously interact with many other agents, e.g., pedestrians walking through crowds, vehicles passing through intersections or merging on highways. Thus, the model discussed in the previous section needs to be extended to consider a general number of agents as well as the spatio-temporal relationships between them. 

\subsection{Modeling a General Number of Agents}

A natural approach to modeling such interactions is to abstract the scene as a spatio-temporal graph (STG) $G = (V, E)$, named so because it represents agents as nodes and their interactions as edges, which evolve in time. An edge $(u, v) \in E$ is present if agent $u$ ``interacts" with agent $v$.
As an input to learned interaction models, spatial proximity is a commonly used proxy for whether two agents may directly interact \cite{AlahiGoelEtAl2016,JainZamirEtAl2016,LeeChoiEtAl2017,DeoTrivedi2018}. Specifically, agents $u$ and $v$ are said to be interacting if $\| \mathbf{p}_u - \mathbf{p}_v \|_2 \leq d$ where $\mathbf{p}_u, \mathbf{p}_v$ are the spatial coordinates of agents $u, v$ in the world, and $d$ is a distance threshold that sets the interaction range of agents. A benefit of abstracting scenes in this way is that it enables any similarly-structured approach to be applied to various environments, and even different problem domains (e.g., modeling human-object interactions in computer vision \cite{JainZamirEtAl2016}), as STGs are general abstractions. \cref{fig:hero} shows an example of a STG abstraction of an autonomous driving scene.

This changes the trajectory forecasting problem from one of modeling agents and their interactions to one of modeling nodes and their edges. The key challenge here is that an agent can have a general number of neighbors which change from one scenario to another. Thus the resulting model needs to be able to handle a general number of inputs for a fixed architecture (since neural network weights have fixed sizes). To do this, one can extend the architecture discussed in \cref{sec:two_agent_cvae} so that it mimics the structure of the scene's STG. In particular, an LSTM is added for each edge that connects to a node (directly modeling edges), with an intermediate aggregation step in order to combine the influence from neighboring nodes of the same type. This is the approach taken in \cite{IvanovicSchmerlingEtAl2018}, which demonstrated that this structure can model neighboring agents' influence.

While this enables one to model a general number of agents, an additional consideration needs to be made for the fact that $V$, the set of agents, and $E$, the set of agent-agent interactions, are \emph{time-varying}. This is especially noticeable in autonomous driving as a vehicle's sensors have limited range. As a result, agents can appear and disappear at every timestep, e.g., due to merging on or off the highway near the ego-vehicle. Even if the number of agents was constant, their interactions are necessarily time-varying as agents' spatial proximity to others changes as they move. Thus, the edge encoding scheme discussed in this subsection needs to be further extended to capture time-varying structure.

\subsection{Modeling Time-varying Interactions}

Introducing time-variance modifies our STG representation from $G = (V, E)$ to $G_t = (V_t, E_t)$. Unfortunately, naively recreating a new STG per timestep and applying the above modeling methodology will be expensive and inefficient as it does not recycle information that may persist over multiple timesteps (e.g., keeping track of which edges are new, established, or recently removed).

Another approach is to introduce a scalar that modulates the outputs of each edge-encoding LSTM depending on how recently the edge was added or removed. This is the approach taken in \cite{IvanovicPavone2019}, where the scalar varies from $0$ to $1$ and acts as an additional weighting factor before the edge's influence is included in the rest of the model. This output re-weighting additionally serves as a low-pass filter so that newly added or removed edges do not wildly swing model outputs from one timestep to another, rejecting high-frequency noise produced by upstream perception systems (e.g., when vehicles dither near the limits of sensor range). A key benefit of this approach is that it is fast to update online, due to the model's stateful representation only needing a few matrix multiplication operations to capture new observations \cite{IvanovicPavone2019}. This is especially important in robotic use cases, which frequently require the ability to run online from streaming data in real-time. We will further discuss runtime considerations in \cref{sec:speed}.

\section{Incorporating Agent Dynamics and Heterogeneous Input Data}\label{sec:dyn_hetero_data}

So far, we have seen how one can probabilistically model a general, time-varying number of interacting agents in a scene. In this section, we will dive deeper into considerations for output structures, specifically those that ensure the feasibility of the output trajectories, as well as methods for including additional sources of information that are commonly available on modern robotic platforms, such as high-definition (HD) maps of the surrounding environment.

\subsection{Producing Dynamically-feasible Outputs}

Common to most approaches in behavior prediction is the eventual need to produce outputs in spatial coordinates as this is where many planning constraints are imposed; indeed a majority of evaluation metrics in the academic behavior prediction literature are defined over spatial coordinates \cite{RudenkoPalmieriEtAl2019}. As a result, most methods either directly produce trajectory samples (e.g., GANs) or utilize intermediate models to convert internal representations to positions with uncertainty (e.g., CVAE-based approaches with decoders that output bivariate GMMs), such as the architecture discussed in the previous sections. However, both of these output structures make it difficult to enforce dynamics constraints, e.g., non-holonomic constraints such as those arising from no slip conditions. The absence of such considerations might lead to predictions that are unrealizable by the underlying actions (e.g., predicting that a car will move sideways). 

To remedy this, we can leverage established ideas in dynamics modeling. When selecting a dynamics model to enforce, one usually finds a trade-off between modeling complexity and computational efficiency. In the case of autonomous driving, however, there is an additional complicating factor in the form of perception requirements. Ideally, agent models would be chosen to best match their semantic type. For example, one would usually model cars on the road using a bicycle model \cite{KongPfeiferEtAl2015}. However, estimating the bicycle model parameters or actions of another vehicle from perception online is very difficult as it requires estimation of the vehicle's center of mass, wheelbase, and front wheel steer angle. A related model which does not have such high estimation requirements is the dynamically-extended unicycle model~\cite{LaValle2006BetterUnicycle}. It strikes a good balance between accuracy (accounting for key vehicular non-holonomic constraints, e.g., no slip constraints) and efficiency (having only four states and two actions), without requiring complex online parameter estimation procedures (one only needs to estimate the vehicle's position and velocity). This choice of dynamics model follows the one made in~\cite{SalzmannIvanovicEtAl2020}, which shows through experiments that such a simplified model is already quite impactful on improving prediction accuracy.

To incorporate such dynamics considerations, one should instead view their learning architecture as producing distributions over an agent's actions rather than positions, and focus on the process of integration from actions to position through the agent's dynamics.
Notably, this scheme can also propagate the model's uncertainty in its generated actions to uncertainty over the resulting positions, especially if the output action uncertainty at each time step has a simple parameterization, e.g., as a Gaussian.
In this case, with linear underlying agent dynamics (e.g., single integrators, frequently used to model pedestrians), the total system dynamics with uncertainty are linear Gaussian. Formally, for a single integrator with actions $\mathbf{u}^{(t)} = \dot{\mathbf{p}}^{(t)}$, the position mean at $t+1$ is $\mathbb{\mu}_{\mathbf{p}}^{(t+1)} = \mu_{\mathbf{p}}^{(t)} + \mu_{\mathbf{u}}^{(t)} \Delta t$, where $\mu_{\mathbf{u}}^{(t)}$ is produced by the learning architecture. In the case of nonlinear dynamics (e.g., unicycle models, used to model vehicles), one can still (approximately) use this uncertainty propagation scheme by linearizing the dynamics about the agent's current state and action.\footnote{Full mean and covariance equations for the single integrator and dynamically-extended unicycle models can be found in the appendix of~\cite{SalzmannIvanovicEtAl2020}.} This dynamics integration scheme is used in \cite{SalzmannIvanovicEtAl2020} and enables the model to produce \emph{analytic} output distributions.

Importantly, even with this additional inclusion of dynamics, no additional data is required for training (e.g., the loss was not amended to be over actions). The model still directly learns to match a dataset's ground truth position, with gradients backpropagated through the agent's dynamics to the rest of the model. Thus, without any extra data, this inclusion of dynamics enables the model to generate explicit action sequences that lead to dynamically-feasible trajectory predictions. Overall, this output scheme is able to guarantee that its trajectory samples are dynamically feasible, in contrast to methods which directly output positions.

\subsection{Incorporating Heterogeneous Data}

Modern robotic systems host a plethora of advanced sensors which produce a wide variety of outputs and data modalities for downstream consumption. However, many current behavior prediction methodologies only make use of the tracked trajectories of other agents as input, neglecting these other sources of information from modern perception systems. 

Notably, HD maps are used by many real-world systems to aid localization as well as inform navigation. Depending on sensor availability and sophistication, maps can range in fidelity from simple binary obstacle maps, i.e., $M \in \{0, 1\}^{H \times W \times 1}$, to multilayered semantic maps, e.g., $M \in \{0, 1\}^{H \times W \times L}$, where each layer $1 \leq \ell \leq L$ indicates areas with specfic semantic type (e.g., road, walkway). 
A major reason for this choice of map format is that it closely resembles images, which also have height, width, and channel dimensions. As a result, Convolutional Neural Networks (CNNs), which are efficient to evaluate online, can be used to incorporate them in behavior prediction models. This is the choice made in \cite{SalzmannIvanovicEtAl2020}, which uses a relatively small CNN to encode local scene context around the agent being modeled.

More generally, one can similarly include further additional information (e.g., raw LIDAR data, camera images, pedestrian skeleton or gaze direction estimates) in the encoder of an architecture by representing it as a vector via an appropriate model and concatenating the resulting output to the encoder's overall scene representation vector.

\section{Experiments and Practical Considerations}\label{sec:expts}

\begin{table}[t]
\vspace{0.85cm}
\centering
\caption{Comparison of our CVAE-based method against GAN-based methods for pedestrian modeling. Bold is best.}
\begin{tabular}{l|cccc}
\toprule
\multicolumn{1}{c|}{\textbf{Method}} & BoN ADE \cite{GuptaJohnsonEtAl2018} & BoN FDE \cite{GuptaJohnsonEtAl2018} & KDE NLL \cite{IvanovicPavone2019} \\ \midrule
S-GAN \cite{GuptaJohnsonEtAl2018}  & $0.58$ & $1.18$ &  $5.80$ \\
S-BiGAT$^*$ \cite{KosarajuSadeghianEtAl2019} & $0.48$ & $1.00$ & - \\
Trajectron++ \cite{SalzmannIvanovicEtAl2020} & $\mathbf{0.21}$ & $\mathbf{0.41}$ & $\mathbf{-1.14}$ \\ 
\bottomrule
\end{tabular}
$^*$Model not public, so we could not evaluate it on the KDE NLL metric. \Tstrut
\label{tab:generative_BoN}
\end{table}

\begin{table}[t]
\vspace{0.5cm}
\centering
\caption{Comparison of our CVAE-based method against others for vehicle modeling. Bold is best.}
\begin{tabular}{l|cccc}
\toprule
\multicolumn{1}{c|}{\textbf{Method}} & FDE@1s & FDE@2s & FDE@3s & FDE@4s \\ \midrule
Const. Velocity & $0.32$ & $0.89$ & $1.70$ & $2.73$\\
S-LSTM$^*$~\cite{AlahiGoelEtAl2016,CasasGulinoEtAl2019} & $0.47$ & - & $1.61$ & - \\
CSP$^*$~\cite{DeoTrivedi2018,CasasGulinoEtAl2019} & $0.46$ & - & $1.50$ & - \\
CAR-Net$^*$~\cite{SadeghianLegrosEtAl2018,CasasGulinoEtAl2019} & $0.38$ & - & $1.35$ & - \\
SpAGNN$^*$~\cite{CasasGulinoEtAl2019} & $0.36$ & - & $1.23$ & - \\
Trajectron++~\cite{SalzmannIvanovicEtAl2020} & $\mathbf{0.07}$ & $\mathbf{0.45}$ & $\mathbf{1.14}$ & $\mathbf{2.20}$\\
\bottomrule
\end{tabular}
$^*$We subtracted detector/tracker errors~\cite{CasasGulinoEtAl2019} as we do not use them. \Tstrut
\vspace{-0.5cm}
\label{tab:nuscenes_results}
\end{table}

In this section, we quantitatively compare the method described in \cref{sec:dyn_hetero_data} against state-of-the-art approaches for the challenging problem of pedestrian and vehicle motion prediction. Additionally, we discuss important implementation considerations for practitioners seeking to employ the approaches presented in this work.

\subsection{Quantitative Performance}

We compare Trajectron++ \cite{SalzmannIvanovicEtAl2020} against Social GAN~\cite{GuptaJohnsonEtAl2018} and Social BiGAT~\cite{KosarajuSadeghianEtAl2019}, all of which use similar RNN-based architectures to model temporal sequences. The approaches are evaluated on the real-world ETH \cite{PellegriniEssEtAl2009} and UCY \cite{LernerChrysanthouEtAl2007} pedestrian datasets, a standard benchmark in the field comprised of challenging multi-human interaction scenarios. We evaluate their performance with the Best-of-$N$ (BoN) Average and Final Displacement Error (ADE and FDE) metrics 
proposed in \cite{GuptaJohnsonEtAl2018} as well as the Kernel Density Estimate-based Negative Log-Likelihood (KDE NLL) proposed in \cite{IvanovicPavone2019}. As can be seen in \cref{tab:generative_BoN}, the CVAE-based Trajectron++ significantly outperforms the others on the three specified metrics. In addition, \cref{tab:nuscenes_results} shows our method's strong vehicle modeling performance against a variety of approaches on the large-scale nuScenes dataset \cite{CaesarBankitiEtAl2019}. Further experiments as well as ablation studies can be found in \cite{SalzmannIvanovicEtAl2020}.
More broadly, the success of phenomenological approaches for large-data regimes has been reflected in modern trajectory forecasting competitions. For instance, all prize winners of the recent ICRA 2020 nuScenes \cite{CaesarBankitiEtAl2019} prediction challenge (one of which is Trajectron++ \cite{SalzmannIvanovicEtAl2020}) are phenomenological, using deep encoder-decoder architectures and leveraging heterogeneous input data in addition to past trajectory history.

\subsection{Latent Space Size}
The size of the latent space (i.e., the number of latent variables) is something not yet discussed in this work. While finding the ``optimal'' size ends up being a hyperparameter search, 
one should generally allocate a latent variable for each high-level behavior or effect they wish to model. In the (common) case where it is difficult to know exactly how many that is (e.g., in driver modeling), one should start high and let the CVAE prune out redundant modes by assigning them very low probabilities. For instance, in each of  \cite{SchmerlingLeungEtAl2018,IvanovicSchmerlingEtAl2018,IvanovicPavone2019,SalzmannIvanovicEtAl2020} we use $25$ latent variables (i.e., $z$ can take $25$ values). Of these, the CVAE only ends up assigning significant probability to a few modes at a time, e.g., moving straight, turning left, turning right, stopping.

In order to determine how many modes are being used, the CVAE's learned weights can be analyzed through the lens of evidential theory, as proposed in \cite{ItkinaIvanovicEtAl2019}. Specifically, one can identify which latent variables have direct evidence that supports their existence, and prune the others without any loss in performance. For instance, \cite{ItkinaIvanovicEtAl2019} found that only $2 - 12$ latent variables have direct evidence in \cite{SalzmannIvanovicEtAl2020} and that the rest can be pruned without any loss in performance.

\subsection{Online Model Runtime}\label{sec:speed}

A key consideration in the development of models for robotic applications is their runtime complexity. 
To achieve real-time performance, one can leverage the stateful representation that spatiotemporal graphs provide. Specifically, the model can be updated online with new information without fully executing a forward pass. For instance, due to our method's use of LSTMs, only the last LSTM cells in the encoder need to be fed the newly-observed data. The rest of the model can then be executed using the updated encoder representation. This update-and-predict scheme is applied in \cite{IvanovicPavone2019,SalzmannIvanovicEtAl2020}, both of which achieve real-time online performance.

\section{Conclusions and Future Work}\label{sec:conclusions_future_work}
We have provided a self-contained tutorial on a CVAE approach to multimodal trajectory prediction for multi-agent interactions. 
Additionally, we have presented a taxonomy of existing state-of-the-art approaches, thereby identifying major methodological considerations and placing our proposed approach in perspective.
In the presence of large amounts of data with potentially heterogeneous data types (e.g., spatial features, images, maps), and non-Markovian settings where future behaviors depend on the history of interaction, our proposed CVAE approach is an attractive model for predicting future human trajectories in multi-agent interactive settings. In particular, our CVAE approach is very flexible, making it easy to include heterogeneous data, account for agent dynamics, and tailor it to different types of model-based planning algorithms.

Future work includes further improvements on the model such as developing ways to make the latent space more interpretable, e.g., through the lens of temporal logic, robustifying against upstream sensor noise, and applying the learned model to generate more realistic simulation agents for testing and validation. 
More broadly, there are still many open questions regarding evaluation metrics and architectural considerations stemming from future integration with downstream planning and control algorithms. These questions are increasingly important now that phenomenological trajectory prediction methods have outweighed others in raw performance, and are targeting deployment on real-world safety-critical robotic systems.

\bibliographystyle{IEEEtran}
\bibliography{IEEEabrv,ASL_papers,main}

\newcommand{\noopsort}[1]{} \newcommand{\printfirst}[2]{#1}
  \newcommand{\singleletter}[1]{#1} \newcommand{\switchargs}[2]{#2#1}
\begin{thebibliography}{10}
\providecommand{\url}[1]{#1}
\csname url@rmstyle\endcsname
\providecommand{\newblock}{\relax}
\providecommand{\bibinfo}[2]{#2}
\providecommand\BIBentrySTDinterwordspacing{\spaceskip=0pt\relax}
\providecommand\BIBentryALTinterwordstretchfactor{4}
\providecommand\BIBentryALTinterwordspacing{\spaceskip=\fontdimen2\font plus
\BIBentryALTinterwordstretchfactor\fontdimen3\font minus
  \fontdimen4\font\relax}
\providecommand\BIBforeignlanguage[2]{{%
\expandafter\ifx\csname l@#1\endcsname\relax
\typeout{** WARNING: IEEEtran.bst: No hyphenation pattern has been}%
\typeout{** loaded for the language `#1'. Using the pattern for}%
\typeout{** the default language instead.}%
\else
\language=\csname l@#1\endcsname
\fi
#2}}

\bibitem{RudenkoPalmieriEtAl2019}
A.~Rudenko, L.~Palmieri, M.~Herman, K.~M. Kitani, D.~M. Gavrila, and K.~O.
  Arras, ``Human motion trajectory prediction: A survey,'' \emph{{Int.\ Journal
  of Robotics Research}}, vol.~39, no.~8, pp. 895--935, 2020.

\bibitem{SohnLeeEtAl2015}
K.~Sohn, H.~Lee, and X.~Yan, ``Learning structured output representation using
  deep conditional generative models,'' in \emph{{Conf.\ on Neural Information
  Processing Systems}}, 2015.

\bibitem{SchmerlingLeungEtAl2018}
E.~Schmerling, K.~Leung, W.~Vollprecht, and M.~Pavone, ``Multimodal
  probabilistic model-based planning for human-robot interaction,'' in
  \emph{{Proc.\ IEEE Conf.\ on Robotics and Automation}}, 2018.

\bibitem{HaanJayaramanEtAl2019}
P.~de~Haan, D.~Jayaraman, and S.~Levine, ``Causal confusion in imitation
  learning,'' in \emph{{Conf.\ on Neural Information Processing Systems}},
  2019.

\bibitem{LeungSchmerlingEtAl2019}
K.~Leung, E.~Schmerling, M.~Zhang, M.~Chen, J.~Talbot, J.~C. Gerdes, and
  M.~Pavone, ``On infusing reachability-based safety assurance within planning
  frameworks for human-robot vehicle interactions,'' \emph{{Int.\ Journal of
  Robotics Research}}, vol.~39, pp. 1326--1345, 2020.

\bibitem{IvanovicSchmerlingEtAl2018}
B.~Ivanovic, E.~Schmerling, K.~Leung, and M.~Pavone, ``Generative modeling of
  multimodal multi-human behavior,'' in \emph{{IEEE/RSJ Int.\ Conf.\ on
  Intelligent Robots \& Systems}}, 2018.

\bibitem{IvanovicPavone2019}
B.~Ivanovic and M.~Pavone, ``The {Trajectron}: Probabilistic multi-agent
  trajectory modeling with dynamic spatiotemporal graphs,'' in \emph{{IEEE
  Int.\ Conf.\ on Computer Vision}}, 2019.

\bibitem{SalzmannIvanovicEtAl2020}
T.~Salzmann, B.~Ivanovic, P.~Chakravarty, and M.~Pavone, ``Trajectron++:
  Dynamically-feasible trajectory forecasting with heterogeneous data,'' in
  \emph{{European Conf.\ on Computer Vision}}, 2020.

\bibitem{HelbingMolnar1995}
D.~Helbing and P.~Moln\'{a}r, ``Social force model for pedestrian dynamics,''
  \emph{{Physical Review E}}, vol.~51, no.~5, pp. 4282--4286, 1995.

\bibitem{TreiberHenneckeEtAl2000}
M.~Treiber, A.~Hennecke, and D.~Helbing, ``Microscopic simulation of congested
  traffic,'' in \emph{Traffic and Granular Flow '99}.\hskip 1em plus 0.5em
  minus 0.4em\relax {Springer Berlin Heidelberg}, 2000.

\bibitem{HelbingFarkasEtAl2000}
D.~Helbing, I.~Farkas, and T.~Vicsek, ``Simulating dynamical features of escape
  panic,'' \emph{{Nature}}, vol. 407, pp. 487--490, 2000.

\bibitem{TreiberHenneckeEtAl2000b}
M.~Treiber, A.~Hennecke, and D.~Helbing, ``Congested traffic states in
  empirical observations and microscopic simulations,'' \emph{{Physical Review
  E}}, vol.~62, no.~2, pp. 1805--1824, 2000.

\bibitem{NikolaidisNathEtAl2017}
S.~Nikolaidis, S.~Nath, A.~D. Procaccia, and S.~Srinivasa, ``Game-theoretic
  modeling of human adaptation in human-robot collaboration,'' in \emph{{IEEE
  Int.\ Conf.\ on Human-Robot Interaction}}, 2017.

\bibitem{WangWangEtAl2019}
M.~Wang, Z.~Wang, J.~Talbot, J.~C. Gerdes, and M.~Schwager, ``Game theoretic
  planning for self-driving cars in competitive scenarios,'' in
  \emph{{Robotics: Science and Systems}}, 2019.

\bibitem{NarayananManogharEtAl2020}
V.~Narayanan, B.~M. Manoghar, V.~S. Dorbala, D.~Manocha, and A.~Bera,
  ``{ProxEmo}: Gait-based emotion learning and multi-view proxemic fusion for
  socially-aware robot navigation,'' in \emph{{IEEE/RSJ Int.\ Conf.\ on
  Intelligent Robots \& Systems}}, 2020.

\bibitem{RandhavaneBeraEtAl2019}
T.~Randhavane, A.~Bera, E.~Kubin, A.~Wang, K.~Gray, and D.~Manocha,
  ``Pedestrian dominance modeling for socially-aware robot navigation,'' in
  \emph{{Proc.\ IEEE Conf.\ on Robotics and Automation}}, 2019.

\bibitem{NgRussell2000}
A.~Ng and S.~Russell, ``Algorithms for inverse reinforcement learning,'' in
  \emph{{Int.\ Conf.\ on Machine Learning}}, 2000.

\bibitem{AbbeelNg2004}
P.~Abbeel and A.~Y. Ng, ``Apprenticeship learning via inverse reinforcement
  learning,'' in \emph{{Int.\ Conf.\ on Machine Learning}}, 2004.

\bibitem{ZiebartMaasEtAl2008}
B.~D. Ziebart, A.~Maas, J.~A. Bagnell, and A.~K. Dey, ``Maximum entropy inverse
  reinforcement learning,'' in \emph{{Proc.\ AAAI Conf.\ on Artificial
  Intelligence}}, 2008.

\bibitem{SadighSastryEtAl2016c}
D.~Sadigh, S.~Sastry, S.~A. Seshia, and A.~D. Dragan, ``Planning for autonomous
  cars that leverage effects on human actions,'' in \emph{{Robotics: Science
  and Systems}}, 2016.

\bibitem{KretzschmarSpiesEtAl2016}
H.~Kretzschmar, M.~Spies, C.~Sprunk, and W.~Burgard, ``Socially compliant
  mobile robot navigation via inverse reinforcement learning,'' \emph{{Int.\
  Journal of Robotics Research}}, vol.~35, no.~11, pp. 1289--1307, 2016.

\bibitem{ChoudhurySwamyEtAl2019}
R.~Choudhury, G.~Swamy, D.~Hadfield-Menell, and A.~D. Dragan, ``On the utility
  of model learning in {HRI},'' in \emph{{IEEE Int.\ Conf.\ on Human-Robot
  Interaction}}, 2019.

\bibitem{AlahiGoelEtAl2016}
A.~Alahi, K.~Goel, V.~Ramanathan, A.~Robicquet, L.~Fei-Fei, and S.~Savarese,
  ``Social {LSTM}: Human trajectory prediction in crowded spaces,'' in
  \emph{{IEEE Conf.\ on Computer Vision and Pattern Recognition}}, 2016.

\bibitem{JainZamirEtAl2016}
A.~Jain, A.~R. Zamir, S.~Savarese, and A.~Saxena, ``Structural-{RNN}: Deep
  learning on spatio-temporal graphs,'' in \emph{{IEEE Conf.\ on Computer
  Vision and Pattern Recognition}}, 2016.

\bibitem{HochreiterSchmidhuber1997}
S.~Hochreiter and J.~Schmidhuber, ``Long short-term memory,'' \emph{{Neural
  Computation}}, 1997.

\bibitem{GoodfellowPouget-AbadieEtAl2014}
I.~Goodfellow, J.~Pouget-Abadie, M.~Mirza, B.~Xu, D.~Warde-Farley, S.~Ozair,
  A.~Courville, and Y.~Bengio, ``Generative adversarial nets,'' in
  \emph{{Conf.\ on Neural Information Processing Systems}}, 2014.

\bibitem{MirzaOsindero2014}
M.~Mirza and S.~Osindero. (2014) Conditional generative adversarial nets.
  {Available at }\url{https://arxiv.org/abs/1411.1784}.

\bibitem{KingmaWelling2013}
D.~P. Kingma and M.~Welling. (2013) Auto-encoding variational bayes. {Available
  at }\url{https://arxiv.org/abs/1312.6114}.

\bibitem{GuptaJohnsonEtAl2018}
A.~Gupta, J.~Johnson, F.~Li, S.~Savarese, and A.~Alahi, ``Social {GAN}:
  Socially acceptable trajectories with generative adversarial networks,'' in
  \emph{{IEEE Conf.\ on Computer Vision and Pattern Recognition}}, 2018.

\bibitem{LeeChoiEtAl2017}
N.~Lee, W.~Choi, P.~Vernaza, C.~B. Choy, P.~H.~S. Torr, and M.~Chandraker,
  ``{DESIRE:} distant future prediction in dynamic scenes with interacting
  agents,'' in \emph{{IEEE Conf.\ on Computer Vision and Pattern Recognition}},
  2017.

\bibitem{DeoTrivedi2018}
M.-F. Deo and J.~Trivedi, ``Multi-modal trajectory prediction of surrounding
  vehicles with maneuver based lstms,'' in \emph{{IEEE Intelligent Vehicles
  Symposium}}, 2018.

\bibitem{KosarajuSadeghianEtAl2019}
V.~Kosaraju, A.~Sadeghian, R.~Mart\'{i}n-Mart\'{i}n, I.~Reid, S.~H.
  Rezatofighi, and S.~Savarese, ``{Social-BiGAT}: Multimodal trajectory
  forecasting using bicycle-{GAN} and graph attention networks,'' in
  \emph{{Conf.\ on Neural Information Processing Systems}}, 2019.

\bibitem{SalimansGoodfellowEtAl2016}
T.~Salimans, I.~Goodfellow, W.~Zaremba, V.~Cheung, A.~Radford, and X.~Chen,
  ``Improved techniques for training {GAN}s,'' in \emph{{Conf.\ on Neural
  Information Processing Systems}}, 2016.

\bibitem{ArjovskyBottou2017}
M.~Arjovsky and L.~Bottou, ``Towards principled methods for training generative
  adversarial networks,'' in \emph{{Int.\ Conf.\ on Learning Representations}},
  2017.

\bibitem{ArjovskyChintalaEtAl2017}
M.~Arjovsky, S.~Chintala, and L.~Bottou, ``Wasserstein generative adversarial
  networks,'' in \emph{{Int.\ Conf.\ on Machine Learning}}, 2017.

\bibitem{JangGuEtAl2017}
E.~Jang, S.~Gu, and B.~Poole, ``Categorial reparameterization with
  gumbel-softmax,'' in \emph{{Int.\ Conf.\ on Learning Representations}}, 2017.

\bibitem{MaddisonMnihEtAl2017}
C.~J. Maddison, A.~Mnih, and Y.~W. Teh, ``The concrete distribution: A
  continuous relaxation of discrete random variables,'' in \emph{{Int.\ Conf.\
  on Learning Representations}}, 2017.

\bibitem{AdelGhahramaniEtAl2018}
T.~Adel, Z.~Ghahramani, and A.~Weller, ``Discovering interpretable
  representations for both deep generative and discriminative models,'' in
  \emph{{Int.\ Conf.\ on Machine Learning}}, 2018.

\bibitem{KongPfeiferEtAl2015}
J.~Kong, M.~Pfeifer, G.~Schildbach, and F.~Borrelli, ``Kinematic and dynamic
  vehicle models for autonomous driving control design,'' in \emph{{IEEE
  Intelligent Vehicles Symposium}}, 2015.

\bibitem{LaValle2006BetterUnicycle}
S.~M. LaValle, ``Better unicycle models,'' in \emph{Planning Algorithms}.\hskip
  1em plus 0.5em minus 0.4em\relax {Cambridge Univ.\ Press}, 2006, pp.
  743--743.

\bibitem{CasasGulinoEtAl2019}
S.~Casas, C.~Gulino, R.~Liao, and R.~Urtasun, ``{SpAGNN}: Spatially-aware graph
  neural networks for relational behavior forecasting from sensor data,'' 2019.

\bibitem{SadeghianLegrosEtAl2018}
A.~Sadeghian, F.~Legros, M.~Voisin, R.~Vesel, A.~Alahi, and S.~Savarese,
  ``{CAR-Net}: Clairvoyant attentive recurrent network,'' in \emph{{European
  Conf.\ on Computer Vision}}, 2018.

\bibitem{PellegriniEssEtAl2009}
S.~Pellegrini, A.~Ess, K.~Schindler, and L.~v. Gool, ``You'll never walk alone:
  Modeling social behavior for multi-target tracking,'' in \emph{{IEEE Int.\
  Conf.\ on Computer Vision}}, 2009.

\bibitem{LernerChrysanthouEtAl2007}
A.~Lerner, Y.~Chrysanthou, and D.~Lischinski, ``Crowds by example,''
  \emph{{Computer Graphics Forum}}, vol.~26, no.~3, pp. 655--664, 2007.

\bibitem{CaesarBankitiEtAl2019}
H.~Caesar, V.~Bankiti, A.~H. Lang, S.~Vora, V.~E. Liong, Q.~Xu, A.~Krishnan,
  Y.~Pan, G.~Baldan, and O.~Beijbom, ``{nuScenes}: A multimodal dataset for
  autonomous driving,'' 2019.

\bibitem{ItkinaIvanovicEtAl2019}
M.~Itkina, B.~Ivanovic, R.~Senanayake, M.~J. Kochenderfer, and M.~Pavone,
  ``Evidential sparsification of multimodal latent spaces in conditional
  variational autoencoders,'' in \emph{{Conf.\ on Neural Information Processing
  Systems}}, 2020, in Press.

\end{thebibliography}

\end{document}